\documentclass[11pt,letterpaper,twocolumn]{article}

\usepackage[paperwidth=8.5in,paperheight=11.0in,
  left=1.0in,right=1.0in,top=1.0in,bottom=1.0in,
  heightrounded]{geometry}

\usepackage{sectsty}
\usepackage{url}
\usepackage{graphicx}
\usepackage{amsmath}
\sectionfont{\normalsize}
\subsectionfont{\normalsize}

\usepackage{titlesec}

\titleformat*{\section}{\large\bfseries}
\titleformat*{\subsection}{\large\itshape}

\titlespacing\section{0pt}{5pt plus 4pt minus 2pt}{0pt plus 2pt minus 2pt}
\titlespacing\subsection{0pt}{5pt plus 4pt minus 2pt}{0pt plus 2pt minus 2pt}
\titlespacing\subsubsection{0pt}{5pt plus 4pt minus 2pt}{0pt plus 2pt minus 2pt}

\usepackage[font=footnotesize,labelfont=bf]{caption}
\usepackage{multicol}

\usepackage[sort&compress]{natbib}
\makeatletter
\renewcommand{\maketitle}{\bgroup\setlength{\parindent}{0pt}
\begin{flushleft}
\textbf{\LARGE \@title}
\vspace*{.2in} \\
\large \@author
\end{flushleft}\egroup
}
\makeatother

\title{Abstraction and Analogy-Making in Artificial Intelligence}
\author{Melanie Mitchell \\ \normalsize Santa Fe Institute, Santa Fe, NM, USA \\
\vspace*{.2in}
Address for correspondence:  Melanie Mitchell, Santa Fe Institute, 1399 Hyde Park Road, Santa Fe, NM 87501 USA.  mm@santafe.edu}
\date{}

\begin{document}
\onecolumn
\maketitle

\textbf{Abstract:} Conceptual abstraction and analogy-making are key abilities underlying humans' abilities to learn, reason, and robustly adapt their knowledge to new domains.  Despite of a long history of research on constructing AI systems with these abilities, no current AI system is anywhere close to a capability of forming humanlike abstractions or analogies. This paper reviews the advantages and limitations of several approaches toward this goal, including symbolic methods, deep learning, and probabilistic program induction.  The paper concludes with several proposals for designing challenge tasks and evaluation measures in order to make quantifiable and generalizable progress in this area.  


\section*{Introduction \label{Introduction}}
\begin{quote} Without concepts there can be no thought, and without analogies there can be no concepts.\\
---D. Hofstadter and E. Sander \cite{Hofstadter2013}
\end{quote}

In their 1955 proposal for the Dartmouth summer AI project, John McCarthy and colleagues wrote, ``An attempt will be made to find how to make machines use language, form abstractions and concepts, solve kinds of problems now reserved for humans, and improve themselves.'' \cite{McCarthy1955}

Now, nearly seven decades later, all of these research topics remain open and actively investigated in the AI community.  While AI has made dramatic progress over the last decade in areas such as computer vision, natural language processing, and robotics, current AI systems almost entirely lack the ability to form  humanlike concepts and abstractions.

For example, while today's computer vision systems can recognize perceptual categories of objects, such as labeling a photo of the Golden Gate as ``a bridge,'' these systems lack the rich conceptual knowledge humans have about these objects---knowledge that enables robust recognition of such objects in a huge variety of contexts. Moreover, humans are able to form abstractions and apply them to novel situations in ways that elude even the best of today's machines. Continuing with the ``bridge'' example, humans can easily understand extended and metaphorical notions such as ``water bridges,'' ``ant bridges,'' ``bridging one's fingers,'' ``bridge of one's nose,'' ``the bridge of a song,'' ``bridging the gender gap,'' ``a bridge loan,'' ``burning one's bridges,'' ``water under the bridge,'' and so on. Indeed, for humans, any perceptual category such as \textit{bridge} is understood via the rich conceptual structure underlying it.  This conceptual structure makes it easy for humans to answer commonsense questions like ``what would happen if you drove across a raised drawbridge?'' or ``what is on each side of a bridge across the gender gap?''  Moreover, conceptual structures in the mind make it easy for humans to \textit{generate} ``bridges'' at different levels of abstraction; for example, imagine yourself forming a bridge from a couch to a coffee table with your leg, or forming a bridge between two notes on your piano with other notes, or bridging differences with your spouse via a conversation.

Most, if not all, human concepts can be abstracted in this way, via analogy (for example, as an interesting exercise, consider the possible abstractions of everyday concepts such as \textit{mirror, shadow, ceiling, driving, and sinking}).  Douglas Hofstadter goes so far as to define concepts by this property: ``a concept is a package of analogies.'' \cite{Hofstadter2001}  Humans' abilities for abstraction and analogy are at the root of many of our most important cognitive capacities, and the lack of these abilities is at least partly responsible for current AI's brittleness and its difficulty in adapting knowledge and representations to new situations.  Today's state-of-the-art AI systems often struggle in transferring what they have learned to situations outside their training regimes, \cite{Mitchell2019a} they make unexpected and unhumanlike errors, \cite{Alcorn2019} and they are vulnerable to ``adversarial examples'' in a very unhumanlike way. \cite{Szegedy2014,Eykholt2018}

Concepts, abstraction, and analogy are foundational areas of study in cognitive psychology.  Psychological theories of concepts have focused on ideas such as core knowledge, \cite{Carey2011,Spelke2007} exemplars and prototypes, \cite{Nosofsky1986,Rosch1999} the ``theory theory'' of concepts, \cite{Gopnik2003} perceptual simulations, \cite{Barsalou2009} experience-based metaphors, \cite{Lakoff2008} and stochastic functions in a probabilistic ``language of thought.'' \cite{Goodman2015}  Some cognitive scientists have postulated that all concepts, even the most abstract ones, correspond to mental models of the world---grounded in perception---that can be simulated to yield predictions and counterfactuals. \cite{Barsalou2009,Fauconnier1997,Johnson2017,Lake2017,Lakoff2012} Others have argued that concept formation, abstraction, and adaptation are all undergirded by processes of analogy---the act of perceiving essential similarities between entities or situations. \cite{Gentner2017,Hofstadter2001,Holyoak1984} Understanding what concepts are---how they are formed, how they can be abstracted and flexibly used in new situations, how they can be composed to produce new concepts---is not only key to a deeper understanding of intelligence, but will be essential for engineering non-brittle AI systems, ones that can abstract, robustly generalize, resist adversarial inputs, and adapt what they have learned to diverse domains and modalities.

The purpose of this paper is to review selected AI research---both old and very recent---on abstraction and analogy-making, to make sense of what this research has yielded for the broader problem of creating machines with these general abilities, and to make some recommendations for the path forward in this field. 

The remainder of the paper is organized as follows.  The next section gives a brief discussion of the role of abstraction and analogy-making in human intelligence, and the need for such abilities in AI. The following sections review several approaches in the AI literature to capturing abstraction and analogy-making abilities, particularly using idealized domains.  The paper concludes with a discussion that appraises the domains and methods discussed here and proposes several steps for making generalizable progress on these issues.

\section*{Abstraction and Analogy-Making in Intelligence \label{AbstractionAnalogyInIntelligence}}
One makes an \textit{conceptual abstraction} one extends that concept to novel situations, including ones that are removed from perceptual entities, as in the examples of ``bridge'' I gave earlier.  The process of abstraction is driven by \textit{analogy}, in which one mentally maps the essence of one situation to a different situation (e.g., such as mapping one's concept of a bridge across a river to a bridge across the gender gap).  We make an analogy every time our current situation reminds us of a past one, or when we respond to a friend's story with ``the same thing happened to me,'' even though ``the same thing'' can be superficially very different.  In fact, the primary way we humans make sense of novel situations is by making analogies to situations we have previously experienced.  Analogies underlie our abilities to flexibly recognize new instances of visual concepts such as ``a ski race''  or ``a protest march.''  Analogies drive our conceptualizations of wholly new situations, such as the novel coronavirus pandemic that erupted in early 2020, in terms of things we know something about---the pandemic has been variously described as a fire, a tsunami, a tornado, a volcano, ``another Katrina,'' or a war. Recognizing abstract styles of art or music is also a feat of analogy-making.  Many scientific insights are based on analogies, such as Darwin's realization that biological competition is analogous to economic competition \cite{Schweber1980} and Von Neumann's analogies between the computer and the brain. \cite{vonNeumann1958} (These are just a few examples; see Ref.~\citenum{Hofstadter2013} for many more examples at all levels of cognition.)

These examples illustrate two important facts: (1) analogy-making is not a rare and exalted form of reasoning, but rather a constant, ubiquitous, and lifelong mode of thinking; (2) analogy is a key aspect not only of reasoning but also of flexible categorization, concept formation, abstraction, and counterfactual inference.    \cite{Gentner2017,Hofstadter2001,Hofstadter2013,Turner1997} In short, analogy is a central mechanism for unlocking meaning from perception.   (See Ref.~\citenum{Bartha2019} for a broad discussion of analogy and its relation to other mechanisms of cognition and reasoning.) 

Analogy-making gives rise to human abilities lacking in even the best current-day AI systems.  Many researchers have pointed out the need for AI systems that are more robust and general---that is, those that can perform sophisticated transfer learning or ``low-shot'' learning, that robustly can figure out how to make sense of novel situations, and that can form and use abstract concepts.  Given this, it seems that analogy-making, and its role in abstraction, is an understudied area in AI that deserves more attention.

In the following sections I review several selected approaches to studying abstraction and analogy making in AI systems, ranging from older symbolic or hybrid approaches such as the structure-mapping approach of Gentner et al.\ and the ``active symbol'' approach of Hofstadter et al., to more recent recent methods that employ deep neural networks and probabilistic program induction.  This is not meant to be an exhaustive survey of this topic, but rather a review of some of the more prominent approaches, ones that I will analyze in the Discussion section below.  Note that I will focus on concepts and analogies that involve multipart situations, rather than the simpler single-relation ``proportional'' analogies such as ``man is to woman as king is to ?'' \cite{Mikolov2013,Sadeghi2015} While the term ``analogy'' often brings to mind proportional analogies like these, the range of human analogy-making is far more interesting, rich, and ubiquitous.

\section*{Symbolic Methods \label{SymbolicMethods}}
Symbolic approaches to abstraction and analogy typically represent input as structured sets of logic statements, with concepts expressed as natural-language words or phrases.  Early examples include Evans' geometric-analogy solver, \cite{Evans1964} Winston's frame-based system for analogy-making between stories, \cite{Winston1980} and Falkenhainer et al.'s Structure-Mapping Engine (SME). \cite{Falkenhainer1986}  In this section I will describe SME, as well as the Active Symbol Architecture of Hofstadter et al., which combines symbolic and subsymbolic elements.

\subsection*{The Structure-Mapping Engine \label{SME}}
The Structure-Mapping Engine (SME) \cite{Falkenhainer1986} is based on Gentner's structure-mapping theory of human analogy-making. \cite{Gentner1983}  In Gentner's theory, analogical mapping of one entity to another depends only on ``syntactic properties of the knowledge representation [describing the entities], and not on the specific content of the domains.'' \cite{Gentner1983} Furthermore, according to the theory, mappings are primarily made between relations rather than object attributes, and analogies are driven primarily by mappings between higher-order relations (the ``systematicity'' principle).

\begin{figure*}[t]
  \centering
  \includegraphics[width=5in]{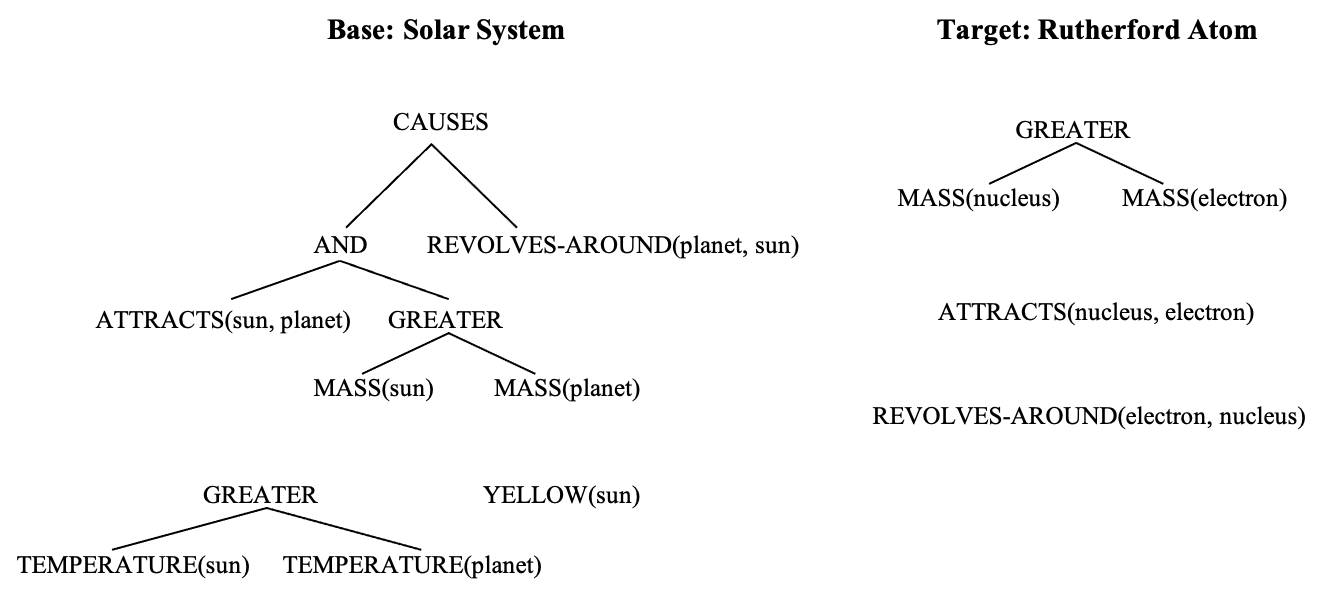}
  \caption{An example of SME's input, in the form of logical propositions.  Adapted from Ref.~\citenum{Falkenhainer1986}.
  \label{SME-Example}}
\end{figure*}

SME's input consists of descriptions of two entities or situations, a \textit{base} and a \textit{target}, where each description consists of a set of logical propositions.  In Figure~\ref{SME-Example}, adapted from one of Falkenhainer et al.'s illustrative examples, the base gives propositions about the solar system and the target gives propositions about the Rutherford atom.

The descriptions in Figure~\ref{SME-Example} are logical statements represented as trees.  The statements include entities (e.g., \textit{planet}), attributes (e.g., \textit{yellow}), first-order relations (e.g., \textit{revolves-around}), and higher-order relations (e.g., \textit{causes}). SME's job is to create a coherent mapping from the base to the target. The program uses a set of ``match rules'' to make candidate pairings between elements of the base and target.  Examples of such rules are: ``If two relations have the same name, then pair them''; ``If two objects play the same role in two already paired relations (i.e., are arguments in the same position), then pair them.''  The program then scores each of the pairings, based on factors such as having the same name, having the same type (e.g., object, attribute, nth-order relationship), and being part of ``systematic structures,'' that is more deeply nested propositions rather than isolated relations.

Once all plausible pairings have been made, the program makes all possible sets of consistent combinations of these pairings, making each set (or ``global match'') as large as possible.  ``Consistency'' here means that each element can match only one other element, and a pair is allowed to be in the global match only if all the arguments of each element are also paired up in the global match.  After all possible global matches have been formed, each is given a score based on the individual pairings it is made up of, the inferences it suggests, and its degree of systematicity (relations that are part of a coherent interconnected system are preferentially mapped over relatively isolated relations).  The output of the system is the set of  possible global matches, ranked by score.

It is important to note that SME is considered to be a model solely of the mapping process of analogy-making; it assumes that the situations to be mapped have already been represented in logical form.  SME has often been used as a mapping module in a system that has other modules for representation, retrieval, and inference. \cite{Blass2016,Dehghani2008,Forbus2005}

One notable example of such a system is the work of Lovett et al.\ \cite{Lovett2017} on solving Ravens Progressive Matrices (RPMs). I'll use this work as an illustrative example, since RPMs have more recently been studied extensively in the deep learning community as a domain for studying abstraction and analogy.  

\begin{figure*}[t]
  \centering
  \includegraphics[width=4in]{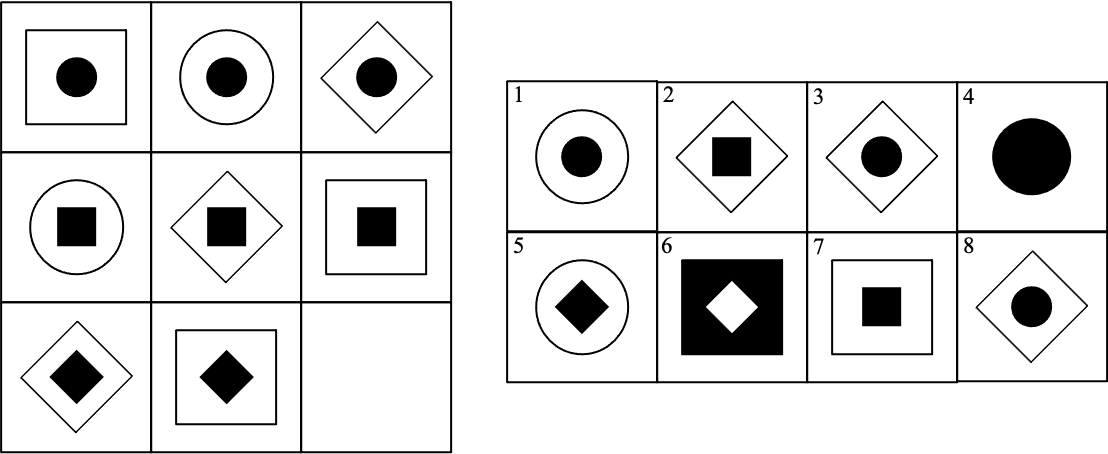}
  \caption{A sample RPM problem.}
  \label{RPM-Problem}
\end{figure*}

RPMs, originally created by John Raven in the 1930s as way to assess ``fluid intelligence,'' have long been given to both children and adults as nonverbal intelligence tests.  Figure~\ref{RPM-Problem} gives a sample problem.  The $3 \times 3$ grid on the left shows patterns of change along the rows and columns; the challenge is to decide which figure from the eight choices on the right fits the blank square. Here the answer is 5.  This example is one of the easier RPM problems; often they involve more than two shapes per square and changes in multiple attributes (shape, color, position, number, etc.)

Lovett et al.\ \cite{Lovett2017} used SME as one component of a system to solve RPM problems.  Their system did not address lower-level vision at all; the input was a vectorized image, pre-segmented (by humans) into objects placed in the grid squares.  The input to SME was a set of predicate-logic descriptions of each square, generated by the CogSketch system. \cite{Forbus2008}  To form these descriptions, CogSketch is programmed with a repertoire of possible object attributes (e.g., size, position, degree of symmetry) and object relations (e.g., inside/outside, intersection, rotation) that it looks for in each box in a given RPM problem.  The system can also create qualitative descriptions of the edges making up objects (e.g., relations such as relative orientation, relative length, etc.) as well as descriptions of object groupings based on proximity and similarity.

These predicate-logic representations are input to SME, which returns the highest-scoring mapping between descriptions of pairs of images in a row.  This mapping is used by the system to determine a higher-level description of the pattern of change across each of the top two rows, from an existing vocabulary of possible changes between corresponding objects (e.g., identity, deformation, shape change, addition/removal).

These higher-level descriptions of the top two rows are themselves mapped by SME to produce a description of more general pattern. SME scores each of the eight possible answers to see which one best completes the third row according to this pattern.  (This brief description leaves out some additional details, such as special-purpose operations for certain kinds of problems.)

Lovett et al.\ ran this system on the Standard Progressive Matrices test. Their system was able to solve 56 out of the 60 problems.  

Lovett et al.'s experiments provide an evaluation of SME as part of a larger system for making analogies in the context of RPMs.  I will say more about the RPM domain in the next section.  

SME captures an important aspect of analogy-making: people tend to prefer systematic analogies involving more abstract concepts rather than less systematic mappings involving more superficial concepts. \cite{Gentner1983}  However, I believe the SME approach is limited in its ability to capture analogy-making more generally, for three main reasons, as I describe below. 

\textbf{Focus on syntax rather than semantics.} The goal of SME is   to give a domain-independent account of analogy-making; thus, true   to its name, SME focuses on mapping the structure or syntax of its   input representations rather than domain-specific semantics.   However, this reliance on syntactic structure requires that representations of situations be cleanly partitioned into objects, attributes, functions, first-order relations, second-order relations, and so on.  The problem is that humans' mental representations of real-world situations are generally not so rigidly categorized.  Building on the ``bridge'' example from the introduction, consider a mapping between the Golden Gate bridge and President Joe Biden's statement that he is a ``bridge'' to a future generation of leaders. \cite{Perkins2020}  Should ``bridge'' be represented as an attribute of the object ``Golden Gate'' (\textit{bridge(GoldenGate)}), an object itself, with its own attributes (e.g., \textit{golden(Bridge)}), or a relation between the areas on either side:   \textit{bridge(GoldenGate, San Francisco, MarinCounty)}? In the latter case, if the Joe Biden ``bridge'' is represented only as a  two-argument relation (\textit{bridge(JoeBiden, FutureGeneration)}), it would not be paired with the three-argument Golden Gate relationship. Or what if ``future generation'' is not a unitary object but an attribute (e.g., \textit{FutureGeneration(Leaders)})?  Then it could not be matched with, say, the ``object'' \textit{MarinCounty}.  These   examples illustrate that real-world situations are not easily reducible to unambiguous propositions that can be mapped via syntax alone.  

The distinctions between ``object,'' ``attribute,'' and ``relation,'' not to mention the order of the relation, depend heavily on context, and people (if they assign such categories at all) have to use them very flexibly, allowing initial classifications to slide if necessary at the drop of a hat.
However, a major tenet of SME is that the mapping process is performed on \textit{existing} representations. This naturally leads into the second issue, the separation of representation-building and mapping.  

\textbf{Separation of representation and mapping processes.} The SME approach separates ``representation-building'' and ``mapping'' into two distinct and independent phases in making an analogy.  The base and target entities are rendered (by a human or another program) into predicate logic descriptions, which is then given to SME to construct mappings between these descriptions.

Some versions of SME enable a limited ``re-representation'' of the   base or target in response to low-scoring mappings. \cite{Forbus1998} Such re-representations might modify certain   predicates (e.g., factoring them into ``subpredicates'') but in the   examples given in the literature, SME almost entirely relies on   another module to build representations before mapping takes place.

In the next section I will discuss a different view, arguing that in human analogy-making, the process of building representations is inextricably intertwined with the mapping process, and that we should adopt this principle in building AI systems that make analogies.  This argument was also made in detail in several previous publications   \cite{Chalmers1992,French2002,Hofstadter1994} and counterarguments were given by the SME authors. \cite{Forbus1998} 

\textbf{Semi-exhaustive search.}  Finally, SME relies on semi-exhaustive search over matchings. The program considers matches between all ``plausible pairings'' (defined by its match rules) to create multiple global matches.  While this is not a problem if representations have been boiled down to a relatively small number of propositions, it is not clear that this semi-exhaustive method will scale well in general.

In short, the structure-mapping theory makes some very useful points about  what features appealing analogies tend to have, but in dealing only with the  mapping process while leaving aside the problem of how situations become  understood and how this process of interpretation interacts with the mapping  process, it leaves out some of the most important aspects of how analogies are  made.

In addition to symbolic approaches such as SME, several hybrid symbolic-connectionist approaches have also been explored, such as the ACME and LISA systems, \cite{Holyoak1989,Hummel1996} neural networks to perform mapping between symbolic representations, \cite{Crouse2020} as well as systems based on cognitive models of memory. \cite{Kokinov2001}

\subsection*{Active Symbol Architecture}
In the 1980s, Hofstadter designed a general architecture for abstract perception and analogy-making that I'll call the ``active symbol architecture,'' based in part on Hofstadter's notion of active symbols in the brain: ``active elements [groups of neurons] which can store information and transmit it and receive it from other active elements'' \cite{Hofstadter1979}---and in part on inspiration from information processing in other complex systems such as ant colonies and cellular metabolism.  

The active symbol architecture was the basis for several AI programs exploring abstract perception and analogy-making, many of which were described in Ref.~\citenum{Hofstadter1995}.  Here I'll focus on Hofstadter and Mitchell's Copycat program. \cite{Hofstadter1994,Mitchell1993}  The name ``Copycat'' is a humorous reference to the idea that the act of making an analogy is akin to being a ``copycat''---that is, understanding one situation and ``doing the same thing'' in a different situation.  A key idea of the Copycat program is that analogy-making should be modeled as a process of abstract perception. Like sensory perception, analogy-making is a process in which one's prior concepts are activated by a situation, either perceived via the senses or in the mind's eye; those activated concepts adapt to the situation at hand and feed back to affect how that situation is perceived.

Hofstadter and Mitchell developed and tested Copycat using the domain of letter-string analogy problems, created by Hofstadter.
\cite{Hofstadter1985}  The following are some examples from this
domain:
\begin{list}{}{}
\item If the string \textbf{abc} changes to the string \textbf{abd}, what does the string \textbf{pqrs} change to?
\item If the string \textbf{abc} changes to the string \textbf{abd}, what does the string \textbf{ppqqrrss} change to?
\item If the string \textbf{abc} changes to the string \textbf{abd}, what does the string \textbf{srqp} change to?
\item If the string \textbf{abc} changes to the string \textbf{abd}, what does the string \textbf{xyz} change to?
\item If the string \textbf{axbxcx} changes to the string \textbf{abc}, what does the string \textbf{pzqzrzsz} change to?
\end{list}

While these analogy problems are idealized ``toy'' problems, they are, similar to Ravens' matrices, designed to capture something of the essence of real-world abstraction and analogy-making.  Each string is an idealized ``situation'' containing objects (e.g., letters or groupings of letters), relations among objects, events (a change from the first to second string), and a requirement for abstraction via what Hofstadter termed \textit{conceptual slippages} \cite{Hofstadter1985} (e.g., the role of ``letter'' in one situation is played by ``group'' in another situation, or the role of ``successsor'' in one situation is played by ``predecessor'' in another situation).

In the Copycat system, the process of analogical mapping between two situations (here, letter strings) is interleaved with the process of building representations of those situations, with continual feedback between these processes.  This is achieved via four interacting components: a \textit{concept network}, which contains the system's prior knowledge in symbolic form; a \textit{workspace}, which serves as a working memory in which representation of and mappings between the input situations takes place; a set of \textit{perceptual agents}, which---competitively and cooperatively---attempt to adapt the system's prior knowledge to the input situations over a series of time steps; and a \textit{temperature} variable, which measures the quality and coherence of the system's representations and mappings at a given time, and which feeds back to control the degree of randomness of the perceptual agents. When the system is far from a solution, the temperature is high, and the perceptual agents' actions are more random; as the system zeroes in on a coherent solution, the temperature falls, and the perceptual agents are more deterministic.

\begin{figure*}[t]
  \centering
  \includegraphics[width=6.5in]{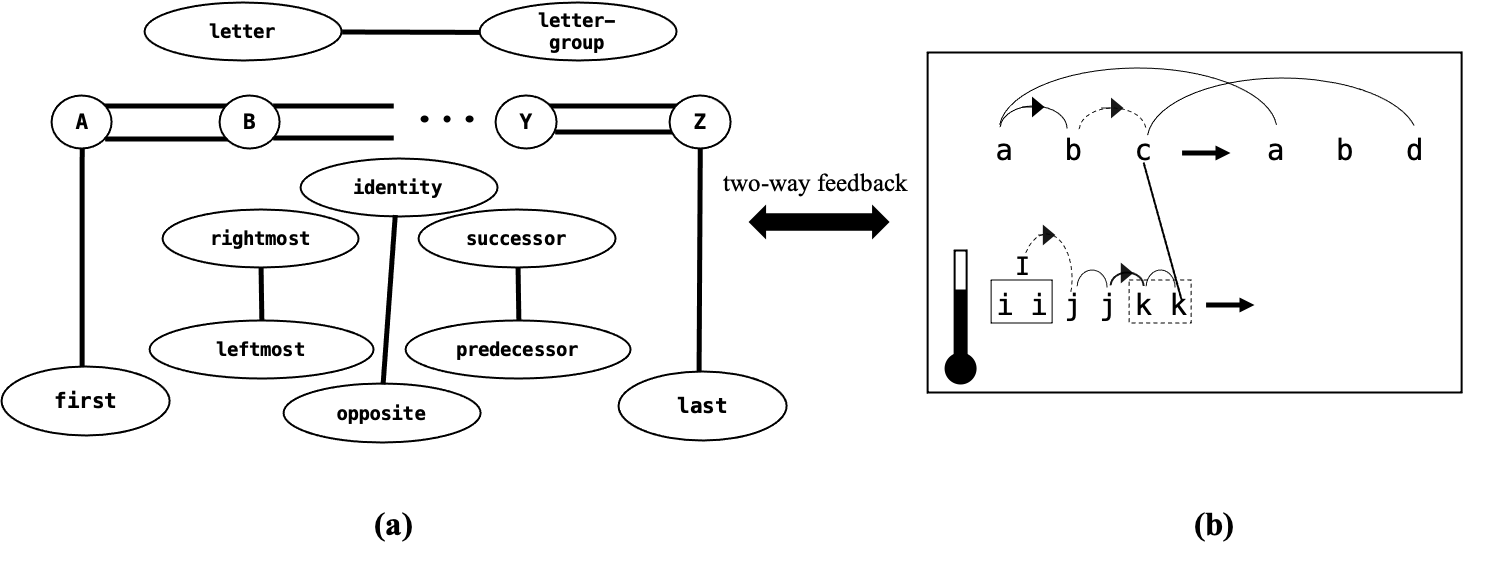}
  \caption{(a) Illustration of part of Copycat’s concept network.  (b) Illustration of Copycat’s workspace, during a run of the program.}
  \label{CopycatArchitecture}
\end{figure*}

\begin{figure*}[t]
  \centering
  \includegraphics[width=6.5in]{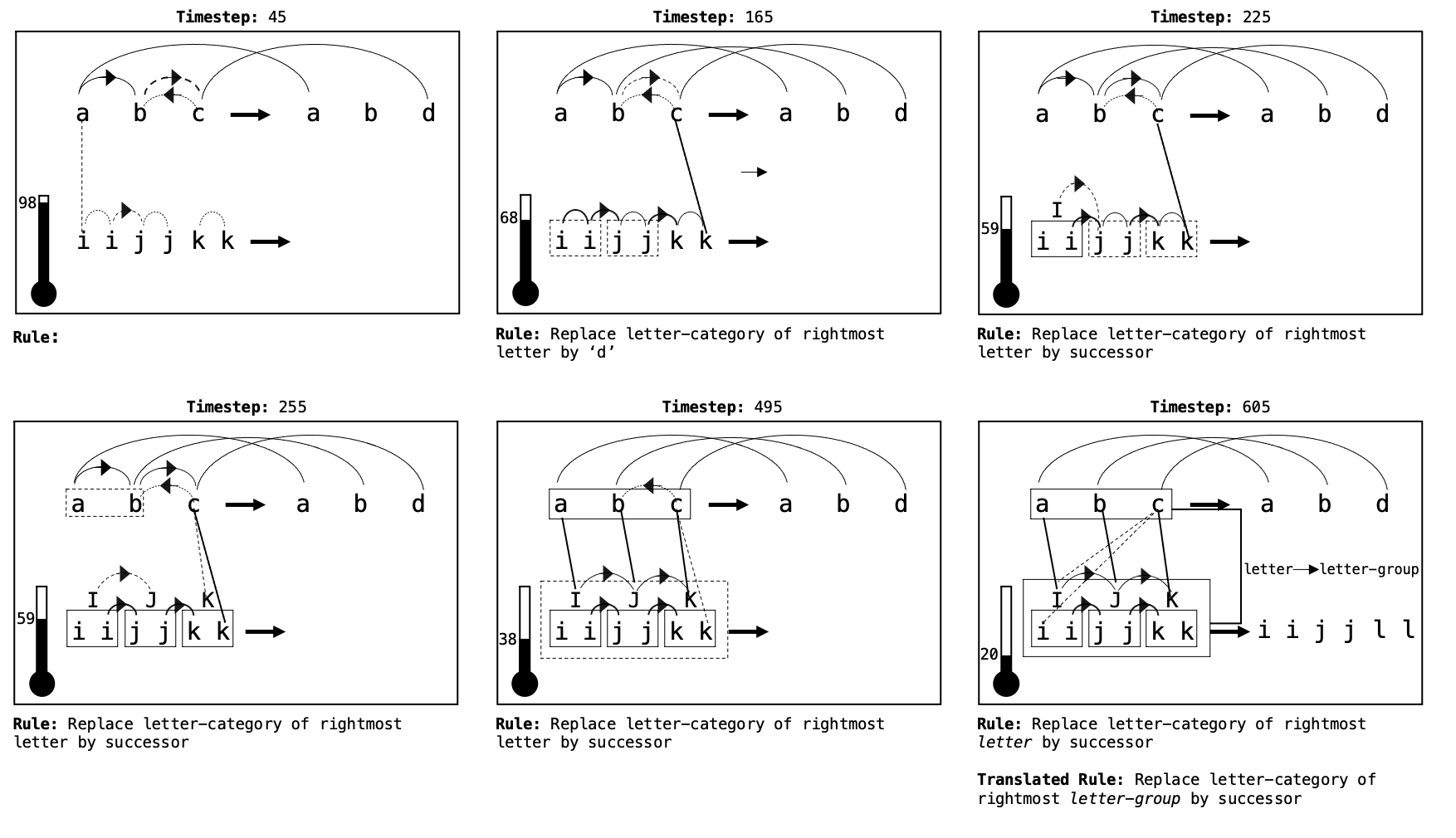}
  \caption{State of the workspace at six different timesteps during a run of Copycat (adapted from Ref.~\citenum{Mitchell1993}).}
  \label{CopycatRun}
\end{figure*}

Figure~\ref{CopycatArchitecture} illustrates the architecture of the Copycat program.  Figure~\ref{CopycatArchitecture}(a) illustrates part of the program's \textit{concept network}, which contains the program’s prior (symbolic) knowledge about the letter-string domain, corresponding to long-term memory.  The concept network models a symbolic ``semantic space,'' in which concepts are nodes (ellipses) and links (lines) between between concepts represent semantic distance, which can change with context during a run of the program. A concept (e.g., \textit{letter-group}) is activated when instances of that concept are discovered in the workspace, and in turn, activated concepts (the program's ``active symbols'') trigger perceptual agents that attempt to discover additional instances. Activation can also spread between conceptual neighbors.  Activation decays over time if not reinforced.

Figure~\ref{CopycatArchitecture}(b) illustrates the program's \textit{workspace}, a short-term memory, inspired by blackboard systems, \cite{Erman1980} in which perceptual agents construct (and sometimes destroy) structures (relations, groupings, correspondences, and rules) that form the program’s current representation of the input situations and the analogy between them, at any given time during a run.  Dashed lines or arcs represent structures with low confidence; solid lines or arcs represent structures with high confidence; the confidence of a structure can change during the run and structures can be destroyed depending on their confidence.  A temperature variable (represented by the thermometer in the bottom right of the workspace) measures the quality of the current structures and feeds back to affect the randomness of the perceptual agents.

Figure~\ref{CopycatRun} gives the state of the workspace at selected timesteps during a run of the program, illustrating how the program constructs representations of, and analogies between, its input situations. The workspace serves as a global blackboard on which agents explore, build, and destroy possible structures. The actions of agents are probabilistic, and depend on the current state of the workspace, concept network, and temperature.  Perceived correspondences between objects in different situations (here letters and letter-groups) lead to conceptual slippages (e.g., \textit{letter} slips to \textit{letter-group}) that give rise to a coherent analogy.  Details of Copycat's operations are described in Ref.~\citenum{Mitchell1993}.

In summary, the Copycat program is an example of Hofstadter's active symbol architecture, in which symbolic concepts become activated via bottom-up perceptions, spread activation to semantically related neighbors, and influence perception in a top-down manner by triggering probabilistic perceptual agents to find instances of the associated concepts in a blackboard-like workspace.  In this way, processing in the system consists of a continual interplay between bottom-up and top-down processes.  A temperature variable controls the degree of randomness in the system and in turn is dynamically determined by the quality of perceptual structures constructed by the system.  Coherent representations of input situations, and analogies between them, result from the perceptual structures constructed by these probabilistic agents.  A central idea underlying the active symbol architecture is that, in analogy-making, the mapping process cannot be separated from the representation-building process---these must be interleaved.  This is a central point of disagreement with the structure-mapping engine approach described in the previous section (see also Ref.~\citenum{Chalmers1992}).

As described in Ref~\citenum{Hofstadter1994}, Copycat's emergent dynamics show a gradual transition from a largely bottom-up (perception-driven), random, and parallel mode of processing---in which many possible representations of the input are explored----to one that is largely top-down (concept-driven), deterministic, and serial.  Copycat does not fit neatly into the traditional AI dichotomy between symbolic and neural systems; rather it incorporates symbolic, subsymbolic, and probabilistic elements.  The architecture resonates with several ideas in psychology, psychophysics, and neuroscience, such as the Global Workspace hypothesis of Baars et al., \cite{Baars2003,Shanahan2006} in which multiple, parallel, specialist processes compete and cooperate for access to a global workspace, and the proposal that visual cortex areas V1 and V2 work as ```active blackboards' that integrate and sustain the result of computations performed in higher areas.'' \cite{Gilbert2007,Roelfsema2016} Copycat also resonates with the idea of neural ``object files'' \cite{Kahneman1992}---temporary and modifiable perceptual structures, created on the fly in working memory, which interact with a permanent network of concepts.  The system's dynamics are also in accord with Treisman's \cite{Treisman1998} notion of perception as a shift from parallel, random, ``pre-attentive'' bottom-up processing and more deterministic, focused, serial, ``attentive'' top-down processing.

Mitchell and Hofstadter showed how Copycat was able to solve a wide selection of letter-string problems; they also described the program's limitations. \cite{Hofstadter1994,Mitchell1993}  The Copycat program inspired numerous other active-symbol-architecture approaches to analogy-making, some of which are described in Ref.~\citenum{Hofstadter1995}, as well as approaches to music cognition, \cite{Nichols2009} image recognition, \cite{Quinn2018} and more general cognitive architectures. \cite{Baars2003}

It bears repeating that Copycat was not meant to model analogy-making on \textit{letter strings} per se.  Rather, the program was meant to illustrate---using the letter-string analogy domain---a domain-independent model of high-level perception and analogy. However, the program has several limitations that need to be overcome to make it a more general model of analogy-making.  For example, Copycat's concept network was manually constructed, not learned; the program illustrated how to adapt pre-existing concepts flexibly to new situations, rather than how to learn new concepts.  Moreover, the program was given a ``source'' and ``target'' situation to compare rather than having to retrieve a relevant situation from memory. Finally, the program's architecture and parameter tuning were complicated and somewhat ad hoc.  Additional research on all of these issues is needed to make active symbol architectures more generally applicable. 

\section*{Deep Learning Approaches \label{DeepLearningApproaches}}
At the other end of the spectrum from symbolic approaches are deep learning approaches, in which knowledge is encoded as high-dimensional vectors and reasoning is the manipulation of these representations via numeric operations in a deep neural network. Moreover, in deep learning, knowledge representation, abstraction, and analogy-making abilities must be learned, typically via large training sets.

Here, I won't survey the substantial literature on pre-deep-learning connectionist modeling of analogy-making (though see, e.g., Refs.~\citenum{Barnden1994,French2002}).  Probably the best-known result of the (post-2010) deep learning era on analogy are the proportional analogies (e.g., ``man is to woman as king is to ?'') that can arise from vector arithmetic on word embeddings. \cite{Mikolov2013}  More recently, numerous papers have been published applying deep learning methods to make analogies between words or simple images \cite{Lu2019,Peyre2019,Sadeghi2015} as well as on simplified Ravens-Progressive-Matrix-like problems. \cite{Hill2019,Webb2020}  In addition, one group demonstrated a deep neural network that can learn to make mappings on symbolic representations (like the ones in Figure~\ref{SME-Example}) that roughly agree with the mappings made by SME. \cite{Crouse2020}

Attempts to get deep learning systems to learn abstract concepts and create analogies are fascinating as ways to explore what these architectures are capable of learning and what kind of reasoning they are able to do.  However, these investigations can be complicated by the amount of data needed to train deep learning systems, as well as by their lack of transparency.  In this paper, as an illustrative example of the promise and pitfalls of deep learning approaches to abstraction and analogy, I'll describe in detail a series of attempts to use deep neural networks on Ravens Progressive Matrices.  This domain has became a popular benchmark in the deep learning community.
    
As I described above, the symbolic Structure Mapping Engine was tested on part of a standard 60-problem RPM test.  However, 60 problems is orders of magnitude too small a dataset to successfully train and test a deep neural net.  In order to apply deep learning to this task, researchers need an automated way of generating a very large number of distinct, well-formed, and challenging RPM problems.

In 2015, Wang and Su \cite{Wang2015b} proposed an RPM-problem-generation method, guided by an earlier analysis of RPM problems. \cite{Carpenter1990}  Their method generated a problem by repeatedly sampling from a fixed set of object, attribute and relation types.  Building on this approach, Barrett et al.\ \cite{Barrett2018} proposed a related method to produce a large dataset of RPM problems, which they called Procedurally Generated Matrices (PGMs).  Each PGM is a set $S$ of triples, where each triple consists of a specific relation, an object type, and an attribute.  The generating process first samples from a fixed set of relation, object, and attribute types to create between one and four triples.  The generating process then samples allowed values for the object types and attributes. All other necessary attributes (e.g., number of objects) are sampled from allowed values that do not induce spurious relationships.  The PGM $S$ is then rendered into pixels.  See  Ref.~\citenum{Barrett2018} for more details on this process, though the paper did not give details on how the eight candidate answers were generated.  

Barrett et al.\ used this generation method to produce their PGM corpus of Raven-style problems. They split the corpus into 1.2M training problems, 20K validation problems, and 200K test problems.  They used these splits to train and test several deep learning methods, including their own novel method, the Wild Relation Network (WReN), which utilized Relation Network modules. \cite{Santoro2017} 

For each network, the input was the eight matrix panels and the eight candidate-answer panels (i.e., 16 grayscale images).  The output was a probability distribution over the eight candidate answers.  Each network was trained to maximize the probability of the correct answer for a given input.  Random guessing would yield 12.5\% accuracy. Among their many experiments, the authors found that their WReN method achieved the best performance of the various networks, an impressive 63\% accuracy on the test problems for the so-called ``neutral split'' (training and test set can contain problems generated with any triples).  Other training/test splits that required extrapolation from a restricted set of triples or attribute values to a different set produced notably poorer performance.  Subsequent to the work of Barrett et al., several other groups developed new deep-learning-based methods that improved state-of-the-art accuracy on the PGM corpus. \cite{Hahne2019,Jahrens2020,Steenbrugge2018,Zheng2019}

In a 2019 paper, Zhang et al.\ \cite{Zhang2019a} questioned whether the PGM dataset was diverse enough to be a good test of abstract reasoning abilities.  They noted that even though the PGM corpus is large (over 1.2 million problems), the PGM generation procedure is limited in the possible kind of problems that can be generated.  Zhang et al.\ noted that ``PGM’s gigantic size and limited diversity might disguise model fitting as a misleading reasoning ability, which is unlikely to generalize to other scenarios.''  In other words, the PGM corpus might make it too easy for learning systems to overfit to its particular, limited types of problems.

To remedy this limitation, Zhang et al.\ devised a new stochastic image-grammar method for generating a more diverse, comprehensive set of Ravens-like problems. They generated a new 70,000-problem corpus called RAVEN---smaller in size than PGM, but more diverse.  Using a split with 42,000 training examples and 14,000 test examples, Zhang et al.\ compared some of the same deep learning methods examined by Barrett et al., plus their own novel ``dynamic residual tree'' method, which dynamically builds a tree-structured computation graph. They found that, whereas Barrett et al.'s WReN method had relatively high accuracy on the PGM dataset, its accuracy on the RAVEN dataset was 15\%, barely above chance.  The highest scoring method, with 60\% accuracy, was a Residual Network \cite{He2016} (ResNet) using features computed by a dynamic residual tree.  The authors also tested humans (college students) on the RAVEN corpus and found human accuracy to be about 84\%.  Other groups were able to improve on the machine-learning accuracy on RAVEN with larger, pretrained networks \cite{Zhuo2020} and contrastive learning, \cite{Zhang2019b} among other novel mechanisms.

\begin{figure}[h]
\centering
\includegraphics[width=2.5in]{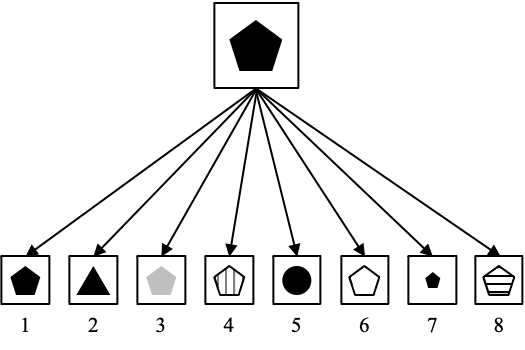}
\caption{Illustration of answer-generation method for RAVEN dataset. Given the correct answer (here, black pentagon at top), each incorrect candidate answer is generated by changing one attribute of the correct answer.  Adapted from from Ref.~\citenum{Hu2021}.}
\label{RAVEN-Answers}
\end{figure}

However, Hu et al.\ \cite{Hu2021} discovered a major flaw in the RAVEN dataset.  Recall that there are eight candidate answers, only one of which is correct. In RAVEN, the seven incorrect answers were generated by randomly modifying a single attribute of the correct answer, as illustrated in Figure~\ref{RAVEN-Answers}.  In the figure, answer \#1 is the correct answer, and each of the other answers differs from it in exactly one attribute.  But this allows a possible shortcut to determining the correct answer: just take the majority vote among the figures' attributes, here, shape, fill-pattern, and size.  Indeed, when Hu et al.\ trained a ResNet only on the eight answer panels in each training example in RAVEN (i.e., leaving out the matrix itself), the ResNet achieved an accuracy of about 90\%, competitive with the state of the art reported in the original RAVEN paper.  This discovery casts doubt on previous claims that deep neural networks that perform well on the RAVEN dataset are actually solving the task that the authors intended them to solve.

Hu et al.\ proposed a more complex sampling method to create what they claimed was an ``unbiased answer set'' for each problem; the result is what they call ``Impartial-RAVEN.''  The authors showed that some deep learning methods reported to do well on RAVEN did much more poorly on Impartial-RAVEN.  Hu et al.\ also described a new method that outperformed other methods.  Impartial-RAVEN (previously called "Balanced-RAVEN") has become the subject of further competition for state-of-the-art accuracy.\cite{Wu2020,Zhang2021} In addition, \cite{Webb2021} demonstrated a mechanism for variable-binding in a neural-network system that performed well on simple abstraction problems including simplified versions of Ravens Progressive Matrices.

In this section I've reviewed some recent work on using deep learning for abstraction and analogy-making in Ravens-like problems.  While this is only one example of work on deep learning in this area, it serves as an illustrative microcosm, and highlights some of the advantages and limitations of such approaches.  On the advantages side, deep learning avoids some of the issues I discussed with SME and Copycat.  Deep learning approaches learn end-to-end from pixels, so avoid relying on built-in knowledge.  Unlike SME, there is no separation of representation-building and mapping, and no semi-exhaustive search over possible matchings.  However, these advantages are at the expense of two major limitations: the need for a huge training set, and the lack of transparency in what has been learned and how it is applied to new problems.  The procedural generation of problems leaves open possibility of biases that allow \textit{shortcuts}---the networks learn ways of performing well on the task that do not require the abstraction and analogy-making mechanisms that we assume humans bring to the task, and that we are trying to capture in machines.  This can be seen in a simple way with biases in the RAVENS set, but such biases can be more subtle and hard to perceive, and have been identified time and again in machine-learning systems. \cite{Funke2020,Geirhos2020,Marasovic2018}  It's not clear that benchmarks such as Balanced-Raven are free from such biases.  This, coupled with the lack of transparency about \textit{how} the networks accomplish their tasks, makes it unclear what these networks have learned, and what features they are basing their decisions on.

All this leads to the conclusion that accuracy alone can be a misleading measure of performance.  I will propose additional evaluation metrics that might be more informative in the discussion section below.  

Even setting aside these problems with biases and lack of transparency, the approach of training such a system with a large set of examples is questionable, given the original intention of RPMs as a way of measuring \textit{general} human intelligence. Humans don't need to train extensively in advance to perform well on RPMs; in fact, doing so would invalidate the test's ability to measure general intelligence.  Rather, the idea is that humans learn how to do successful abstraction and analogy-making in the world, and intelligence tests such as Raven's are meant to be a measure of these general skills.  In contrast, the neural networks that do well on Raven's test after extensive training are unable to transfer any of these skills to any other task. It's not clear that such approaches bring us any closer to the goal of giving machines general abilities for abstraction and analogy-making.

A potentially more promising (though less explored) set of approaches to abstraction uses \textit{meta-learning} \cite{Wang2020}---enabling a neural network to adapt to new tasks by training it on a training set of \textit{tasks} rather than only on examples from a single task.  Several meta-learning approaches for few-shot learning have been evaluated on Lake et al.'s Omniglot dataset, \cite{Lake2015a} among other benchmarks \cite{Finn2017,Ren2018,Santoro2016,Snell2017}. Meta-learning methods produced substantially improved generalization abilities for agents in a grounded-language task in a simple simulated environment, \cite{Hill2021} and improved few-shot learning accuracy in a simplified version of Bongard problems.\cite{Nie2020}  However, at the time of this writing, meta-learning approaches have not yet been explored for other abstraction and analogy domains. Another proposed approach is that of ``meta-mapping'' \cite{Lampinen2020} which directly maps a representation of one task to a representation of a related task.

\section*{Probabilistic Program Induction}
So far I have covered methods illustrating symbolic and deep learning approaches.  In this section I'll describe a different family of methods for concept learning and analogy-making: probabilistic program induction methods, in which a space of possible programs is defined (often via a program grammar) and a probability distribution is computed over this space with respect to a given task. Solving the task amounts to sampling from the space of possible programs guided by this probability distribution.  Here, the task is concept induction, and a concept is identified with a program.  I will illustrate probabilistic program induction by describing its application to two task domains: (1) recognizing and generating handwritten characters, and (2) solving Bongard problems.   As I will discuss, program-induction methods can combine both symbolic and neural-network representations. 

\subsection*{Recognizing and Generating Handwritten Characters} 
Lake et al. \cite{Lake2015a} applied probabilistic program induction to the task of learning visual concepts of handwritten characters from 50 different writing systems---the ``Omniglot'' task.  The goal was to study \textit{one-shot learning}---learning a concept from a single example.  Here, learning a concept means seeing one example of a handwritten character and being able to not only \textit{recognize} new examples of this character, but also being able to \textit{generate} new examples.  In this work, a concept is represented as a program that generates examples of the concept.

\begin{figure}[h]
\centering
\includegraphics[width=3in]{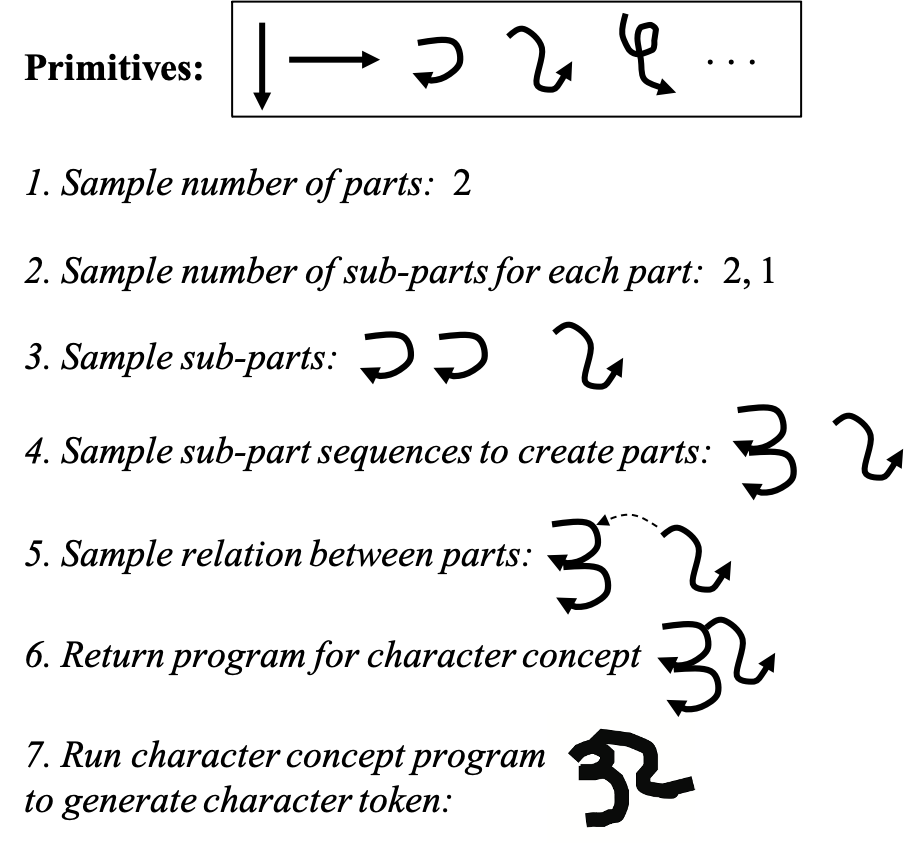}
\caption{Illustration of the generative model used by Lake et al.\ for generating Omniglot characters, given a set of primitive pen strokes.  To generate a new character, first sample the number of parts; then sample the number of sub-parts that will make up each part; then sample these sub-parts from the primitives set; then sample possible sequences of   the sub-parts to make a part; then sample relations between parts (from a library of possible relations). This sequence of samples is the program for generating the character concept.  To generate a token of the character concept (i.e., a rendered character), run the character-concept program with motor noise added to selected points.
Adapted from Ref.~\citenum{Lake2015a}.}
\label{Omniglot}
\end{figure}

In Omniglot, a program representing a concept consists of a hierarchical character-generation procedure.  A character consists of a number of \textit{parts} (complete pen strokes), each of which itself is made of \textit{sub-parts} (movements separated by pauses of the pen).  The generation procedure consists of sampling from a hierarchy of generative models describing the probabilities of the sub-parts, parts, and relationships among them. See Figure~\ref{Omniglot} for an example.

How is this hierarchy of generative models learned?  In Ref.~\citenum{Lake2015a} the authors describe a system that obtains prior knowledge from a training set of human-drawn characters spanning 30 alphabets (taken from the Omniglot dataset).  The training set included the human-drawn pen strokes for each character, as well as the final character images. The learning system collected a library of primitive pen strokes and learned probability distributions over features of these pen strokes (e.g., starting positions).  It also learned transition probabilities between strokes and probabilities over types of relations among the strokes, among other probabilities. All in all, this collection of probabilities is the prior knowledge of the system.

The system used this prior knowledge in a Bayesian framework for the tasks of one-shot classification and generation of characters not contained in the training set.  (The authors also explored related tasks but I won't cover those here.)  In the one-shot classification task, the system was presented with a single test image $I^{(t)}$ of a new character of class $t$, along with 20 distinct characters $I^{(c)}$ in the same alphabet produced by human drawers. Only one of the 20 was the same class as $I^{(t)}$.  The task was to choose that character from the 20 choices.

The Omniglot system computed an approximation to the probability $P(I^{(t)}|I^{(c)})$ for each of the 20 $I^{(c)}$, and chose the $I^{(c)}$ that yielded the highest probability.  By Bayes rule, $$P(I^{(t)}|I^{(c)}) \propto P(I^{(c)}|I^{(t)}) P(I^{(t)}).$$ The term $P(I^{(t)})$ is approximated by a probabilistic search method to generate a program to represent $I^{(t)}$ of the form shown in Figure~\ref{Omniglot}; the prior probabilities learned from the original training set can be used to approximate $P(I^{(t)})$. The term $P(I^{(c)}|I^{(t)})$ can be approximated by attempts to ``refit'' the program representing $I^{(t)}$ to $I^{(c)}$.  See Ref.~\citenum{Lake2015b} for details.

In the experiments reported by Lake et al., \cite{Lake2015a} the Omniglot system's performance on one-shot classification matched or exceeded that of the humans tested on this task.  

The one-shot generation task was, given an example image $I^{(c)}$ of a hand-drawn character of class c, to generate a new image of a character of class $c$.  The Omniglot system did this by first searching to find a program representing $I^{(c)}$ as above, and then running this program as in Figure~\ref{Omniglot} to generate a new example.  Lake et al. enlisted human judges to compare the characters generated by their system with those generated under the same one-shot conditions by humans---what they called a ``visual Turing Test.''  The judges were typically not able to distinguish between machine-generated and human-generated characters.  

Feinman and Lake \cite{Feinman2021} proposed an interesting ``neuro-symbolic'' extension of Lake et al.'s system that integrated neural networks with probabilistic programs to learn generative models of handwritten characters. 

\subsection*{Solving Bongard Problems}
A second example of recent work on probabilistic program induction is Depeweg et al.'s system for solving Bongard programs. \cite{Depeweg2018}  Here the idea is to induce a \textit{rule} rather than a runnable program, but the general idea is the same.

\begin{figure*}[t]
\centering
\includegraphics[width=4in]{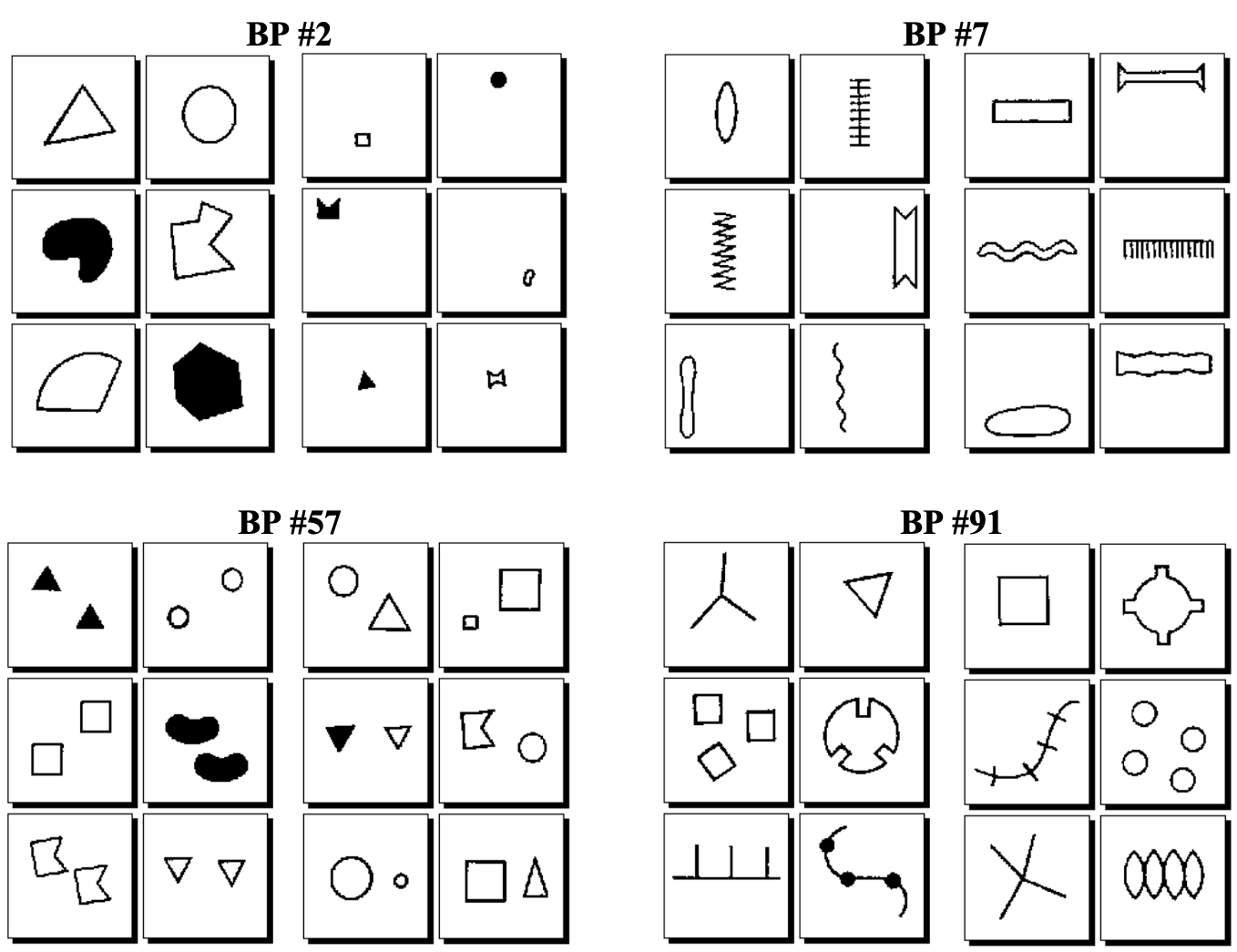}
\caption{Four sample Bongard Problems (adapted from Foundalis' website. \cite{FoundalisWebSite})}
\label{BongardExamples}
\end{figure*}

Bongard problems are visual concept recognition tasks, first presented in M. Bongard's 1970 book \textit{Pattern Recognition}. \cite{Bongard1970}  Figure~\ref{BongardExamples} shows four sample problems from Bongard's collection. For each problem the task is to identify the concept that distinguishes the set of six frames on the left from the set of six on the right (e.g., \textit{large vs.\ small} or \textit{three vs.\ four}). Often the concept is simple to express, but is represented quite abstractly in the figures (e.g., BP \#91). 

Bongard devised 100 such problems as a challenge for artificial intelligence systems.  Harry Foundalis created a web site \cite{FoundalisWebSite} to make available the original 100 problems plus many additional Bongard-like problems created by other people. It is notable that in the decades since Bongard's book was published, no artificial vision system has come close to solving all of Bongard's original problems.

Douglas Hofstadter has written extensively about Bongard problems, starting with his 1979 book \textit{G\"odel, Escher, Bach: an Eternal Golden Braid}, \cite{Hofstadter1979} in which he sketched a rough architecture for solving them.  (Notably, this sketch became the basis for the Active Symbol Architecture that I described above.)

In contrast to the attention garnered in the AI community by Ravens Progressive Matrices, Bongard problems have seen less attention.  In a 1996 paper, Saito and Nakano \cite{Saito1996} showed that an inductive logic programming approach could solve 41 of the original 100 problems, starting not from the raw pixels but from logic formulas created by humans to represent each problem.  In his 2006 PhD dissertation, Foundalis \cite{Foundalis2006} described the construction of a ``cognitive architecture inspired by Bongard's problems,'' whose input was raw pixels of the 12 frames and whose output was an English phrase describing one or both sides of the problem.  Foundalis's architecture was meant to model human concept induction in general rather than Bongard problems specifically, and was able to reliably solve about 10 problems.  

\begin{figure*}[t]
\centering
\includegraphics[width=5in]{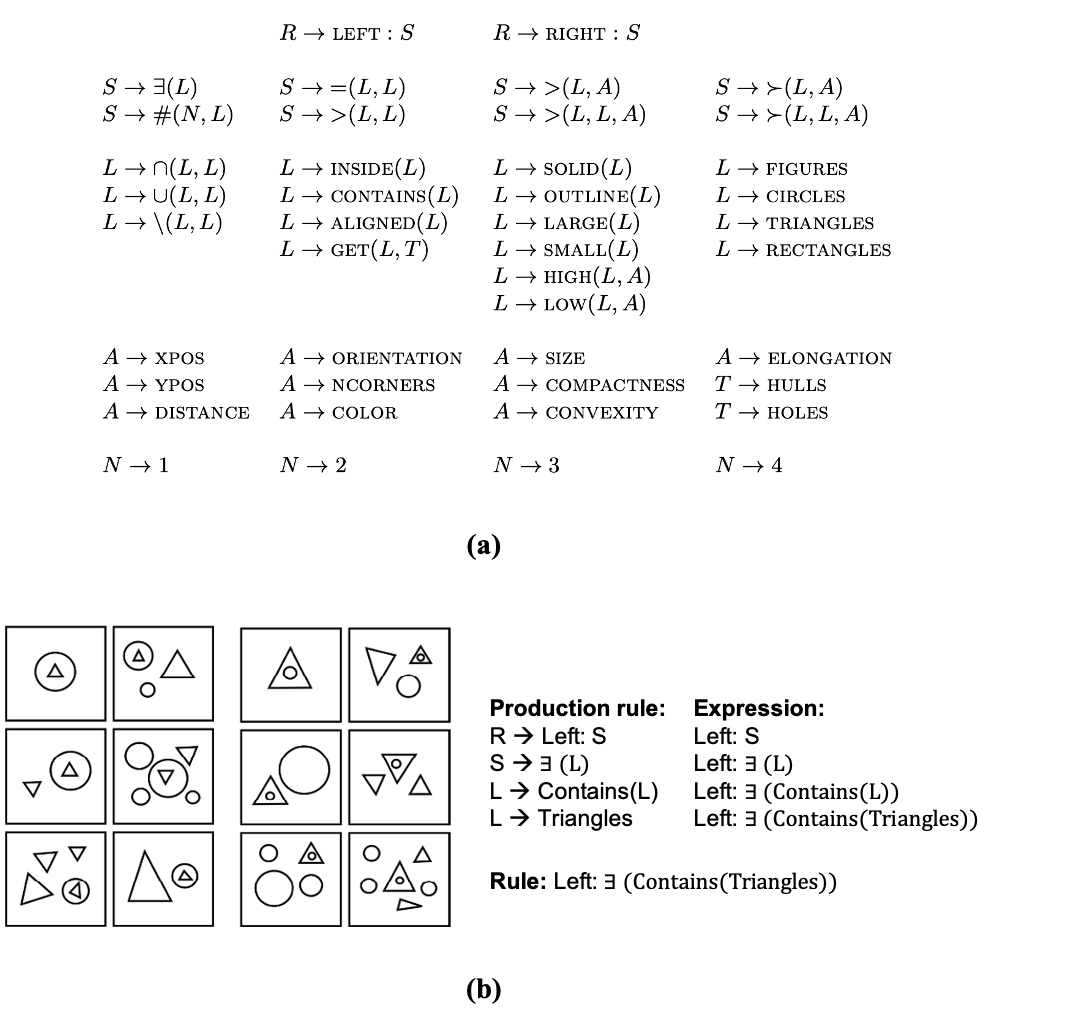}
\caption{(a) Human-designed grammar for rule induction. (b) Sample Bongard problem, with rule derivation and corresponding rule expression ("the left side has objects that contain triangles.") (Adapted from Ref.~\citenum{Depeweg2018}.)}
\label{BongardRuleInduction}
\end{figure*}


Here I'll describe Depeweg et al.'s \cite{Depeweg2018} probabilistic rule-induction approach, which was inspired by the Omniglot system described above.  Depeweg et al.'s system's task was to input the raw pixels of the 12 frames, and to output a rule $R$---in a logic-like language---that is true of all frames on one side (left or right) and none of the frames on the other side.  (Note that this differs from Bongard's original task, which was to output English-language expressions contrasting the left and right sets of boxes.) 

In Depeweg et al.'s system, the space of possible rules is given by a human-designed grammar, given in Figure~\ref{BongardRuleInduction}(a).  A rule can be derived from this grammar by choosing one of the sides (right or left), starting with the start symbol $S$ and probabilistically choosing one of the possible expansions of that symbol (e.g., $S \rightarrow \exists (L)$), then probabilistically choosing possible expansions of the symbols in that expression (e.g., $L \rightarrow Contains(L)$) and so on until all the variables have been expanded.

Given a Bongard problem, the goal is to search the possible space of rules to find the most probable rule.  The authors define a probability distribution $P(R | E,G)$ over possible rules.  Here $R$ is a rule, $E = (E_{LEFT}, E_{RIGHT})$ is the set of 12 input frames, and $G$ is the grammar.  By Bayes rule, this probability distribution can be factored into prior probability $P(R | G)$ and likelihood $P(E | R)$:  
$$ P(R|E,G) \propto P(E|R)P(R|G).$$
The authors define the likelihood $P(E|R)$ as equal to 1 if two conditions are met, and equal to zero otherwise.  The conditions are (1) $R$ is true of all the frames on one side (left or right) and none of the frames on the other side, and (2) the frames in $E$ are ``informative'' about $R$, meaning that the frames must contain the objects or relationships that are mentioned in the rule.  For example, the rule ``there exist triangles or squares inside circles'' is true of all the frames on the left side of Figure~\ref{BongardRuleInduction}(b), but since no squares actually appear, these frames are not considered to be \textit{informative} about the rule, and the rule would be given zero likelihood.  The prior probability term $P(R|G)$ is defined by the authors in a way that favors shorter rules, along with some other structural properties of rules (see Ref.~\citenum{Depeweg2018} for details).

Given the raw pixels of a Bongard problem, Depeweg et al.'s system applies simple image processing operations to extract objects, attributes, and relationships in each frame.  The system was given a small repertoire of object, attribute, and relation types that it was able to extract.  The authors note, ``Our aim was not to build a general vision system, instead we decided to focus on these visual shapes, properties and relations that appear in many of the Bongard problems and hence seem to be likely candidates for a natural vocabulary from which relevant visual concepts can be built.''  The limitations of the system's repertoire means that their system is able to deal with only a subset of 39 out of the 100 original problems. 
Once the image processing has been completed on the input Bongard problem, the system searches for a rule by sampling (using a form of the Metropolis-Hastings algorithm) from possible rules under the probability distribution described above.  The system is allowed 300,000 samples to find a compatible rule for each Bongard problem.  Of the 39 problems the grammar could deal with, the system was able to find compatible rules for 35 of them.  

\subsection*{Summary of Program Induction Approaches \label{ProgramInductionSummary}}
The Omniglot and Bongard-problem examples illustrate the promise and challenges of probabilistic program induction approaches.  Framing concept learning as the task of generating a \textit{program} enables many of the advantages of programming in general, including flexible abstraction, reusability, modularity, and interpretability.  Bayesian inference methods seamlessly combine prior knowledge and preferences with likelihoods, and enable powerful sampling methods.  Such advantages have fueled strong interest and progress in probabilistic program induction in recent years, including combining program induction with neural networks, reinforcement learning, and other methods,  \cite{Chen2019,Ellis2018,Ganin2018,Ellis2020}, as well as  incorporating program induction with methods inspired by neuroscience \cite{Gredilla2019} and psychology. \cite{Evans2021,Rule2020}   The work of L\'{a}zaro-Gredilla et al.\cite{Gredilla2019} is notable in that it combines probabilistic program induction with a neurally inspired model of visual perception and action in a physical robot.  The work of Evans et al.\cite{Evans2021} notably focuses on unsupervised program induction constrained by domain-independent properties of cognition.

Many challenges remain to make probabilistic program induction a more general-purpose AI method for concept learning, abstraction, and analogy.  These methods currently need substantial built-in knowledge, structured by humans, in the form of the program primitives and grammar (the ``domain-specific language'') for a given problem. Moreover, these methods require humans to define prior probability and likelihood distributions over possible programs, which is not always a straightforward task. Perhaps most important, solving a given task can require an enormous amount of search in the space of possible programs, which currently limits scaling up such program induction methods to more complex problems (though some of the hybrid methods cited above are focused on dealing with this combinatorial explosion of possibilities.)  Finally, it remains to be seen whether the concept-as-program notion allows for the flexibility, extensibility, and analogy-making abilities that are the hallmark of human concepts.

\section*{Abstraction and Reasoning Corpus \label{ARC}}
This section does not describe a new AI method for abstraction and analogy, but rather a new promising benchmark---the Abstraction and Reasoning Corpus (ARC)---for evaluating these abilities.  ARC was developed by Fran\c{c}ois Chollet as part of a project on how to measure intelligence in AI systems. \cite{Chollet2019}

\begin{figure*}[t]
  \centering
  \includegraphics[width=4in]{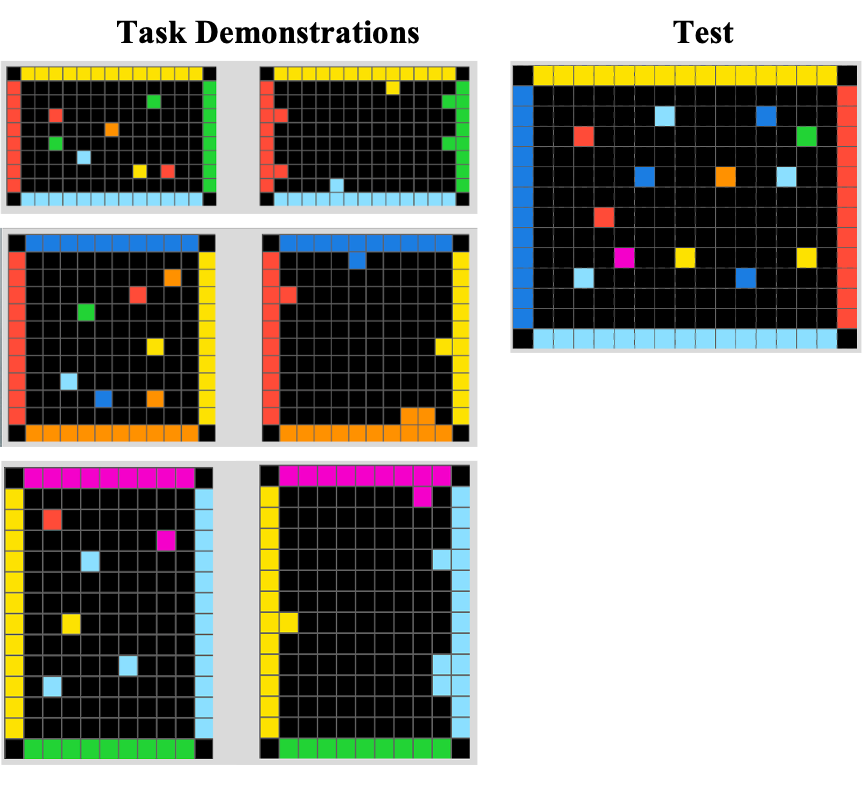}
  \caption{A sample ARC task (adapted from Chollet \cite{ARCGithub2019}). Best viewed in color.}
  \label{ARC-Example}
\end{figure*}


ARC is a collection of visual analogy ``tasks.'' Figure~\ref{ARC-Example} gives a sample task from the corpus.  The left side presents three ``task demonstrations''; each of these consists of two grids with colored boxes.  In each demonstration, you can think of the first (left) grid as ``transforming'' into the second (right) grid.  The right side of Figure~\ref{ARC-Example} presents a ``test''; the task is to transform this grid analogously to the demonstrations.  This task could be thought of as one of ``few-shot'' learning---the demonstrations on the left are the training examples, and the grid on the right is a test example.

The ARC domain features visual analogies between idealized ``situations'' that can express abstract concepts in unlimited variations. In this way it exhibits some of the combined advantages of the other idealized domains I've discussed in this paper, such as Copycat's letter-string analogies, Bongard problems, and Ravens Progressive Matrices. 

Chollet manually designed 1,000 ARC tasks, with the motivation of enabling a fair comparison of ``general intelligence'' between AI systems and humans---a comparison that does not involve language or other acquired human knowledge.  Instead, ARC tasks are meant to rely only on the innate core knowledge systems proposed by Spelke, \cite{Spelke2007} which include intuitive knowledge about objects, agents and their goals, numerosity, and basic spatial-temporal concepts.

An important aspect of ARC is its relatively small size.  Chollet made 400 tasks publicly available and reserved  600 tasks as a ``hidden'' evaluation set.  In 2020 the Kaggle website hosted a three-month ``Abstraction and Reasoning Challenge'' in which researchers were encouraged to submit programs to be evaluated on the hidden evaluation set.  The best-performing submissions had about 20\% accuracy on a top-3 metric (three answers were allowed per task, and if one or more was correct, the task was considered to be solved). However, none of the submissions used an approach that was likely to be generalizable. \cite{CholletPersonalCommunication2020}  Thus the ARC challenge remains wide open.

\section*{Discussion: How to Make Progress in AI on Abstraction and Analogy \label{Discussion}}
In the sections above I have described several diverse AI approaches to abstraction and analogy, including symbolic methods, deep learning, and probabilistic program induction.  These approaches each have their own advantages and limitations. 

Symbolic systems such as SME can be explicitly programmed with prior knowledge (in symbolic form) as well as with important heuristics of abstraction and analogy making, such Gentner's systematicity principle.  Symbolic representation methods such as predicate logic or semantic networks offer unambiguous variables and variable binding, clear type-token distinctions, and explicit measures of conceptual similarity and conceptual abstraction, among other facilities associated with human reasoning. These systems also have the advantage of interpretability, since their ``reasoning'' on a given problem is readable in symbolic form.  However, representations that focus on the syntax of logic-like representations can suffer from brittleness; moreover symbolic approaches often require humans to create and structure substantial prior knowledge, and these systems often rely on semi-exhaustive search.  There have been several interesting \textit{neuro-symbolic} approaches that implement symbolic-like behavior in neural networks (e.g., see Refs.~\citenum{Doumas2020,Mao2019,Smolensky1990}), but these remain limited in their generality and can suffer from some of the limitations of symbolic systems discussed above.

Active symbol architectures, such as the Copycat program, were claimed to address some of the limitations of purely symbolic methods \cite{Chalmers1992,Hofstadter1995} by enabling the system to actively build its representation of a situation in a workspace, via a continual interaction between bottom-up and top-down information processing, avoiding any kind of exhaustive search.  However, Copycat and other examples of active symbol architectures remain dependent on prior knowledge provided and structured by humans, and these systems as yet have no mechanism for learning new permanent concepts.   

Deep learning systems do not require the pre-programmed structured knowledge of symbolic systems; they are able to learn from essentially raw data.  However, they require large training corpora, as well as significant re-training (and often re-tuning of hyperparameters) for each new task.  In addition, they are susceptible to learning statistical shortcuts rather than the actual concepts that humans intend, and their lack of intepretability often makes it difficult to ascertain the extent to which they are actually performing abstraction or analogy.  Moreover, if the goal is to imbue machines with general humanlike abstraction abilities, it doesn't make sense to have to train them on tens of thousands of examples, since the essence of abstraction and analogy is few-shot learning. While some meta-learning systems have shown interesting performance on certain few-shot-learning tasks, their overall generality and robustness still needs to be demonstrated.

Probabilistic program induction, by framing concept learning as the task of generating a \textit{program}, enables many of the advantages of programming in general, including flexible abstraction, reusability, modularity, and interpretability. Moreover, Bayesian inference methods enable the combination of prior knowledge and preferences with likelihoods, and a probabilistic approach enables powerful sampling methods.  However, like symbolic approaches, current program induction approaches require significant human-engineered prior knowledge in the form of a domain-specific language.  And the more expressive the language, the more daunting the combinatorial search problem these methods face.

Stepping back from any individual approach, it is difficult to assess how much general progress has been made on AI methods for abstraction and analogy, since each AI system is developed for and evaluated on a particular domain.  Moreover, as I described in the sections above, these evaluations have largely relied on the system's accuracy on a particular set of test problems.  What's missing are assessments based on generality across diverse domains, as well as the robustness of a given system to factors such as noise, variations on an abstract concept, or scaling of task complexity.  In order to make further progress, we need to rethink how we choose or design domains and what evaluation criteria and evaluation processes we use.

The following are my own recommendations for research and evaluation methods to make quantifiable and generalizable progress in developing AI systems for abstraction and analogy. 

\textbf{Focus on idealized domains.}
AI researchers have used diverse idealized domains for developing and evaluating systems to perform abstraction and analogy, including Ravens Progressive Matrices, letter-string analogy problems, the Omniglot challenge, and Bongard problems, among others (e.g., more recently, new idealized visual domains have been proposed in Refs.~\citenum{Gredilla2019,Nie2020}.) While some approaches have used natural image- or language-based domains, \cite{Andonian2020,Blass2016,Mikolov2013,Lu2019,Peyre2019,Sadeghi2015}
there are advantages to using idealized non-linguistic domains.  In idealized domains, it is possible to be explicit about what prior knowledge is assumed, as opposed to the open-ended knowledge requirements of human language and imagery.  Real-world image- or language-based tasks have much richer meanings to humans than to the machines processing this data; by avoiding such tasks we can avoid anthropomorphizing and overestimating what an AI system has actually achieved.  There are risks with idealized domains, since it is not always clear that these domains capture the kind of real-world phenomena we want to model, but as in other sciences, the first step to progress is to isolate the phenomenon we are studying, in as idealized a form as possible.   While I believe that important general cognitive abilities can be developed using these idealized challenges, others have argued that such abilities can be enabled only by exposing systems to much richer data or experience.\cite{Hill2020}

\textbf{Focus on ``core knowledge.''} Chollet has suggested that challenge domains for assessing AI's ``intelligence'' should rely only on human ``core knowledge'' rather than acquired knowledge such as language.\cite{Chollet2019}  Spelke and colleagues \cite{Spelke2007} have proposed that human cognition is founded on of a set of four core knowledge systems, which include objects and intuitive physics; agents and goal-directedness; numbers and elementary arithmetic; and spatial geometry of the environment (which includes relational concepts such as ``in front of'' or ``contains).  The idealized domains I have discussed above---especially the ARC domain---mostly rely only on such core knowledge.  (Note that in the letter-string analogy domain, letters of the alphabet are used as idealized representatives of objects that can be related to other objects in specific ways; knowledge of language, such as how letters are used in words, or the visual characteristics of letters, are outside of the idealized domain. \cite{Hofstadter1994}) Restricting a domain's required prior knowledge to such concepts makes it possible to fairly compare performance among AI systems as well as between AI systems and humans. Indeed, I would argue that unless we create AI systems that can master such core, non-linguistic knowledge, we have little hope of creating anything like human-level AI.

\textbf{Evaluate systems across multiple domains.} All of the domains I have discussed here capture interesting facets of abstraction and analogy-making, including the recognition of abstract similarity, conceptual extrapolation, and adaptation of knowledge to novel situations.  However, the common practice of focusing on a single domain limits progress.  AI research focusing exclusively on any one domain has the risk of “overfitting” to that domain.  I believe that the research community needs to adopt a diverse suite of challenge domains on which systems can be evaluated for generality; such a strategy has a better chance to develop truly general and robust approaches.   This is the strategy taken by the natural language processing community, for example, with the GLUE and SuperGLUE benchmarks, \cite{Wang2018,Wang2019} though these benchmarks focus on tasks associated with large training sets and fixed test sets, which have allowed for successful solutions based on
``shortcuts.'' \cite{Geirhos2020,Linzen2020,McCoy2019}   For that reason I endorse a focus on tasks that do not allow large amounts of specific training data, as I detail below.  

\textbf{Focus on tasks that require little or no training.} Since abstraction and analogy are at core concerned with flexibly mapping one's knowledge to new situations, it makes little sense to have to extensively train a system to deal with each new abstraction or analogy problem.  Thus I believe the research community should focus on tasks that do not require extensive training on examples from the domain itself.  This echoes Chollet's criteria for such tasks: ``It should not be possible to `buy' performance on the benchmark by sampling unlimited training data.  The benchmark should avoid tasks for which new data can be generated at will.  It should be, in effect, a game for which it is not possible to practice in advance of the evaluation session.'' \cite{Chollet2019}  When  humans make abstractions or analogies, it can be argued that the ``training'' for such abilities is the process of developing core concepts, much of which takes place in early childhood. \cite{Carey2011,Mandler1992,Spelke2007} Similarly, AI systems that can solve problems in  idealized domains should require ``training'' only  on the core concepts required in each domain.  Rather than training and testing on the same task, enabling machines to have general abstraction abilities will require that they learn core concepts and then adapt that knowledge to a multitude of different tasks, without being trained specifically for any one of them.

\textbf{Include generative tasks.} Some idealized domains offer \textit{discriminative} tasks---e.g., Raven's Progressive Matrices, in which the solver chooses from a set of candidate answers.  Others, such as the letter-string analogies or ARC, are \textit{generative}: the solver has to generate their own answer.  Generative tasks are likely more resistant to shortcut learning than discriminative tasks, and systems that generate answers are in many cases more interpretable.  Most importantly, if the space of problems is diverse enough, having to generate answers forces a deeper understanding of the task.  

\textbf{Evaluate systems on ``hidden,'' human-curated, changing sets of problems.} 
It is important that problems to be used for evaluation are not be made available to the developers of AI systems that will be evaluated on these problems.  Furthermore, these evaluation sets should not remain fixed for a long period of time, since systems can indirectly ``overfit'' to a fixed set of evaluation problems, even when they are hidden.  In addition, the evaluation problems should be created and curated by humans, rather than relying on automatic generation by algorithms; I described above how procedurally generated problems can  unintentionally allow shortcut solutions; they can also allow a system to reverse-engineer the generating algorithm instead of being able to solve the problems in a general way. \cite{Chollet2019}   Of course, human-generated problems can also allow for shortcut solutions; thus evaluation sets need to be carefully curated to avoid shortcut solutions as much as possible, for example via adversarial filtering methods. \cite{Sakaguchi2020}

\textbf{Evaluate systems on their robustness, not simply their accuracy.} Like other research in AI, methods for abstraction and analogy are often evaluated on their accuracy on a set of test problems from a given domain.  However, as I discussed above, measuring accuracy on a fixed set of test problems does not reveal possible shortcuts---strategies a system takes to solve problems that don't reflect the actual general abilities that the evaluation is meant to test.  In order to make progress on general abstraction and analogy abilities, in addition to evaluating systems across multiple domains, we need to evaluate them along multiple dimensions of robustness.  For example, the evaluation benchmarks should feature various kinds of challenges in order to measure a system's robustness to ``noise'' and other irrelevant distractions, and to variations in a given concept (e.g., if one example tests recognition of the abstract concept ``monotonically increasing,'' other examples should test variations of this concept with different degrees of abstraction). Finally, the evaluation problems should also test a system's ability to scale to more complex examples of a given concept (e.g., if a system is able to recognize ``monotonically increasing'' with small number of elements, it should also be tested on the same concept with a larger number of elements, or with more complex elements). 

\section*{Conclusion}
In this paper I have argued that humanlike abstraction and analogy-making abilities will be key to constructing more general and trustworthy AI systems, ones that can learn from a small number of examples, robustly generalize, and reliably adapt their knowledge to diverse domains and modalities. I have reviewed several approaches to building systems with these abilities, including symbolic and ``active symbol'' approaches,  deep learning, and probabilistic program induction. I have discussed  advantages and current limitations of each of these approaches, and argued that it remains difficult to assess progress in this area due to the lack of evaluation methods addressing generality, robustness, and scaling abilities.   Finally, I proposed 
several steps towards making quantifiable and generalizable progress, by designing appropriate challenge suites and evaluation methods.  

The quest for machines that can make abstractions and analogies is as old as the AI field itself, but the problem remains almost completely open.  I hope that this paper will help spur renewed interest and attention in the AI community to understanding these core abilities which form the foundations of general intelligence.  

\section*{Acknowledgments}
This material is based upon work supported by the National Science Foundation under Grant No.\ 2020103.  Any opinions, findings, and conclusions or recommendations expressed in this material are those of the author and do not necessarily reflect the views of the National Science Foundation. This work was also supported by the Santa Fe Institute.  I am grateful to Marianna Bolognesi, Andrew Burt, Richard Evans, Ross Gayler, Brenden Lake, Adam Santoro, Wai Keen Vong, and two anonymous reviewers for comments on an earlier version of the manuscript. 

\bibliography{ANYAS.bib}

\end{document}